\documentclass[runningheads]{llncs}
\usepackage[T1]{fontenc}
\usepackage{graphicx}
\usepackage{amsmath,amssymb,amsfonts}
\usepackage{tcolorbox}
\usepackage{subcaption}
\usepackage{booktabs}
\usepackage{array}
\usepackage{hyperref}

\newcolumntype{R}{>{\raggedleft\arraybackslash}p{1cm}}
\begin{document}
\title{Hierarchical Text Classification\\
       Using Black Box Large Language Models}
%
%
\iftrue
\author{Kosuke Yoshimura \and Hisashi Kashima}
\authorrunning{K. Yoshimura and H. Kashima}
\institute{Kyoto University, Kyoto, Japan\\
\email{yoshimura.kosuke.42e@st.kyoto-u.ac.jp \\ kashima@i.kyoto-u.ac.jp}
}
\else
\author{Anonymous Author(s)}
\authorrunning{Anonymous Author et al.}
\institute{Anonymous Institute}
\fi
\maketitle              
\begin{abstract}
Hierarchical Text Classification (HTC) aims to assign texts to structured label hierarchies; however, it faces challenges due to data scarcity and model complexity. This study explores the feasibility of using black box Large Language Models (LLMs) accessed via APIs for HTC, as an alternative to traditional machine learning methods that require extensive labeled data and computational resources. We evaluate three prompting strategies—Direct Leaf Label Prediction (DL), Direct Hierarchical Label Prediction (DH), and Top-down Multi-step Hierarchical Label Prediction (TMH)—in both zero-shot and few-shot settings, comparing the accuracy and cost-effectiveness of these strategies.
Experiments on two datasets show that a few-shot setting consistently improves classification accuracy compared to a zero-shot setting. While a traditional machine learning model achieves high accuracy on a dataset with a shallow hierarchy, LLMs, especially DH strategy, tend to outperform the machine learning model on a dataset with a deeper hierarchy. API costs increase significantly due to the higher input tokens required for deeper label hierarchies on DH strategy. These results emphasize the trade-off between accuracy improvement and the computational cost of prompt strategy. These findings highlight the potential of black box LLMs for HTC while underscoring the need to carefully select a prompt strategy to balance performance and cost.


\end{abstract}
\section{Introduction}
Hierarchical Text Classification~(HTC) is a text classification problem in which labels are structured hierarchically. The goal is to classify a given text into one or more appropriate labels from a predefined hierarchical label set~\cite{hdltex}.
With the rapid expansion of digital content, vast amounts of textual information are generated daily. Manually organizing and retrieving relevant information from such an overwhelming volume is infeasible.
HTC plays a crucial role in systematically categorizing documents, enabling efficient information retrieval. Applications of HTC include the classification of medical texts~\cite{Gargiulo2019,MOSKOVITCH2006177}, academic articles~\cite{hdltex}, and user reviews on e-commerce platforms~\cite{Bhambhoria2023}.
However, HTC is inherently challenging due to the large number of candidate labels, often ranging from hundred to thousand. This leads to two key issues: (1) \textbf{Data scarcity}—as the number of labels increases, labeled training data for each category becomes sparse, making it difficult to train robust models; and (2) \textbf{Model complexity}—when dealing with a vast label space, traditional machine learning and deep learning models often suffer from overfitting or underfitting due to insufficient data per label.

Recent advancements in Large Language Models~(LLMs)~\cite{gpt3,llama2} have demonstrated their ability to perform zero-shot and few-shot learning across various tasks~\cite{zeroshottextclassifiers,tabllm}. Given the challenges of HTC, LLMs offer a promising solution by enabling classification with minimal labeled data while eliminating the need for training complex models from scratch.
There are two main ways to utilize LLMs. The first method involves setting up and using publicly available large language models or self-trained language models on one's computational resources, which we refer to as \textbf{white box LLMs}. The second method involves using APIs or GUIs provided by LLM providers, referred to as \textbf{black box LLMs}. While white box LLMs allow fine-tuning and direct access to model parameters, their deployment requires substantial computational resources, making them costly.
In contrast, black box LLMs can be used via APIs, requiring no computational overhead for training, making them a more practical solution for lightweight HTC implementation.

\begin{figure}[t]
    \centering
    \includegraphics[width=0.75\textwidth]{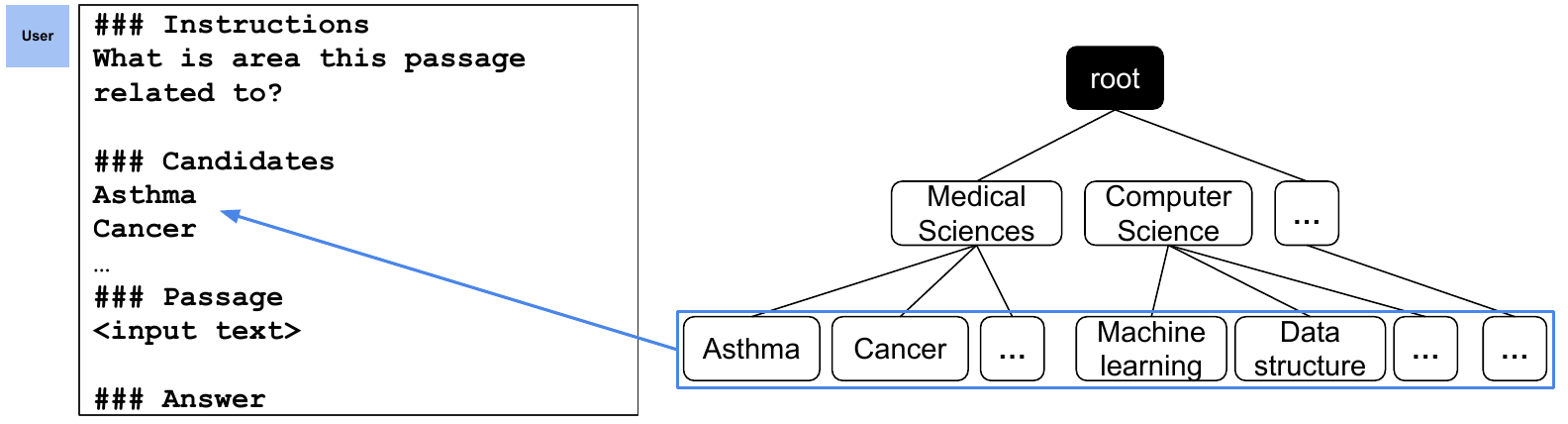}
    \caption{Direct Leaf Label Prediction Strategy.}
    \label{fig:teaser_dl}
\end{figure}

\begin{figure}[t]
    \centering
    \includegraphics[width=0.75\textwidth]{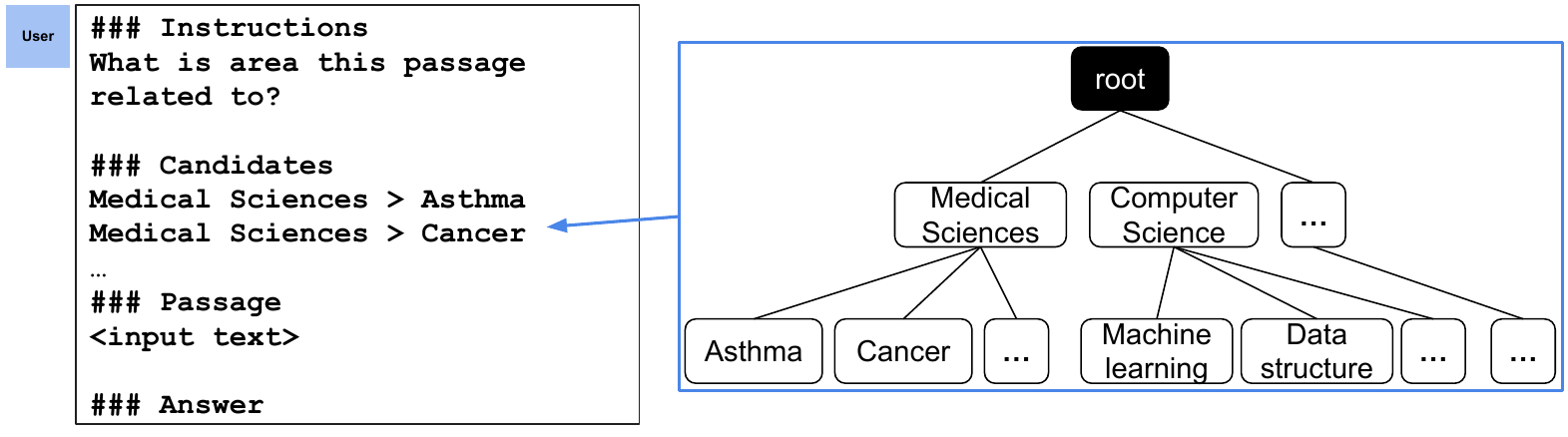}
    \caption{Direct Hierarchical Label Prediction Strategy.}
    \label{fig:teaser_dh}
\end{figure}

\begin{figure}[t]
    \centering
    \includegraphics[width=0.7\textwidth]{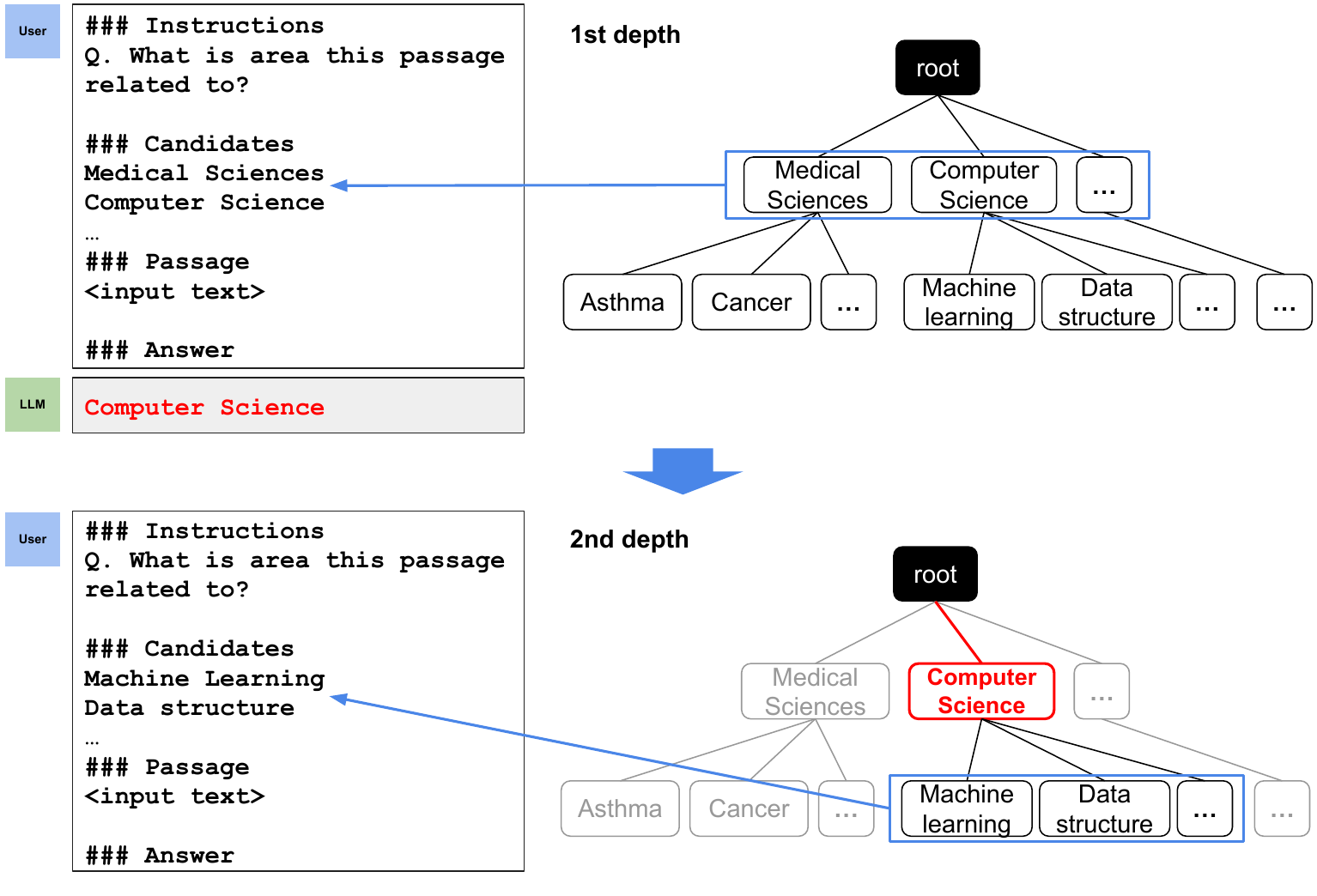}
    \caption{Top-down Multi-step Hierarchical Label Prediction Strategy. In this figure, since LLM selects \textbf{``Computer Science''} at the 1st depth, this approach provides only the child nodes of \textbf{``Computer Science''} as the candidate labels to prompt text at the 2nd depth.}
    \label{fig:teaser_tmh}
\end{figure}

This study investigates the feasibility of using black box LLMs for HTC by adapting typical machine learning and deep learning HTC strategies into prompting techniques. We evaluate multiple prompting strategies in both few-shot and zero-shot settings, comparing their classification accuracy and cost.
We conducted experiments in both few-shot and zero-shot settings to compare the accuracy and cost of hierarchical text classification based on different prompting strategies, thereby identifying the potential of black box LLMs for this task.


The contributions of this paper are summarized as follows:
\begin{itemize}
    \item We apply three prompt strategies to adapt the typical approaches used in solving HTC to LLMs.
    \item  We conduct a comprehensive evaluation of these prompting strategies on real-world datasets, assessing both accuracy and cost. 
    \item  We compare few-shot and zero-shot performance to highlight the effectiveness and limitations of Black Box LLMs for HTC.
\end{itemize}

\section{Related Works}
This section accounts for previous research on hierarchical text classification, applications of LLMs, and hierarchical text classification with LLMs, respectively.
In particular, for hierarchical text classification with LLMs, we also discuss the differences between this and existing studies.

\subsection{Hierarchical Text Classification}
Hierarchical Text Classification is a problem setup in which text is given as input, and one or more appropriate labels are selected from a set of candidate labels with a hierarchical structure.
In classical machine learning, methods have been proposed to create features from the input text and apply classification models~(e.g., SVM) specific to Hierarchical Classification~\cite{hsvm,bsvm}.

Kowsari et al. proposed an approach that uses an appropriate deep learning architecture for each hierarchy to address the problem that traditional multi-class classification approaches lose accuracy as the number of labels increases~\cite{hdltex}. This approach trains a model to estimate child labels with data conditioned on the parent labels for each hierarchy. However, this approach does not overcome the fact that a large amount of training data is required to train the deep-learning models.

Gargiulo et al. noted that PubMed datasets with deep hierarchies are not given all labels from root to leaf when labeling experts. To solve this issue, they proposed Hierarchical Label Set Expansion~(HLSE) to complement the relation between parent node and child node~\cite{Gargiulo2019}.

Wang et al. proposed HPT~(Hierarchy-aware Prompt Tuning) for hierarchical text classification, which integrates label hierarchy information into dynamic virtual templates and hierarchy-aware label words, thereby bridging the gap between conventional prompt tuning and the training tasks of pre-trained language models~(PLMs)~\cite{hpt}. Furthermore, by introducing a zero-bounded multi-label cross-entropy loss, it effectively addressed issues of label imbalance and low-resource scenarios~\cite{hpt}.
Ji et al. propose HierVerb, a multi-verbalizer framework for few-shot hierarchical text classification that directly embeds hierarchical information into layer-specific verbalizers~\cite{hierverb}. By integrating a hierarchy-aware constraint chain and flat hierarchical contrastive loss, HierVerb effectively leverages pre-trained language model knowledge, achieving significant performance gains over graph encoder-based methods~\cite{hierverb}.
Both HPT and HierVerb are not based on large language models; instead, they leverage language models such as BERT.

Several previously proposed techniques for text hierarchical classification rely on complex implementations and huge training data to optimize their machine learning models. While these approaches have demonstrated strong performance, their resource-intensive nature contrasts sharply with our objective of achieving hierarchical classification through a streamlined, lightweight implementation.

\subsection{Applications of LLMs}
In recent years, many providers have started to make their own trained LLMs available to the public, triggered by OpenAI's ChatGPT.
This has led to many studies attempting to use LLMs to solve various problems and tasks~\cite{tabllm,llmrec,zeroshottextclassifiers}.

Wang et al. proposes and validates using LLMs as zero-shot text classifiers. 
Their research is similar to this study but differs in two key aspects. Firstly, they have not evaluated text classification that considers hierarchical structures. Secondly, they focus solely on zero-shot scenarios. In contrast, this study examines prompting strategies for hierarchical text classification and considers the few-shot case.

\subsection{Hierarchical Text Classification with LLMs}
Several studies have explored the application of LLMs to HTC~\cite{Bhambhoria2023,teleclass,chen-etal-2024-retrieval-style,10825412}. This section highlights key differences between these approaches and our research.

Bhambhoria et al. proposed a model that combines LLMs and Entailment Predictors by converting a hierarchical classification task into a long-tail prediction task~\cite{Bhambhoria2023}. Bhambhoria et al. The proposed combined method performs better than using LLMs and entailment predictors individually through experiments.
Their research does not explicitly compare strategies that consider the hierarchical structure of labels within prompts. Moreover, while we focus on prompting strategies, they focus on a framework that combines LLMs with entailment-contradiction predictors, which differs from our problem setting.

Zhang et al. proposed a hierarchical text classification method called TELEClass~\cite{teleclass}. TELEClass achieved high-performance classification model training using LLMs for annotation and expanding the taxonomy. While their research focuses on the zero-shot setting, our research also addresses few-shot prompting. Moreover, their research aims to fine-tune a pre-trained model using only the corpus and label names to achieve high-precision hierarchical text classification, whereas our research aims to elucidate the differences in accuracy and cost for each prompting strategy in hierarchical text classification.

Chen et al. propose a retrieval-based in-context learning (ICL) approach for few-shot HTC, utilizing a retrieval database and pre-training with hierarchical classification and contrastive learning~\cite{chen-etal-2024-retrieval-style}. Their method requires task-specific training and database construction. In contrast, our work explores Black Box LLMs, which do not require retrieval or fine-tuning but rely solely on prompting strategies. We analyze the accuracy-cost trade-offs in few-shot and zero-shot settings, demonstrating the efficiency and limitations of Black Box LLMs for HTC.

Schmidt et al. focus on zero-shot hierarchical text classification, leveraging LLMs with hierarchical label structures to improve classification performance~\cite{10825412}. Their study explores prompt-based classification but does not conduct few-shot experiments, limiting their evaluation to scenarios where no labeled examples are available.

In contrast, our study systematically examines both zero-shot and few-shot settings, providing a more comprehensive analysis of prompting strategies in HTC. By incorporating few-shot experiments, we assess how providing a small number of labeled examples impacts accuracy and cost, offering insights into the trade-offs between supervision levels and classification performance. This distinction highlights the broader applicability of our work, particularly in practical settings where a limited amount of labeled data is available.

\section{Problem Settings}

This study aims to elucidate a method for achieving low-cost and high-accuracy hierarchical classification (HTC) with a lightweight implementation.
Thus, the goal is to solve the HTC problem by using only Black Box LLMs for inference, adopting a zero-shot or few-shot setting.

Given a text $X$ and a black box large language model $f$ that we can only use through API calls. We assume that we cannot train or fine-tune the large language model $f$ in this setting. The goal is to assign more accurate labels from the candidate label set $Y$ corresponding to this input text $X$ using the large-scale language model $f$ by devising prompting strategies. This candidate label set $Y$ has a hierarchical structure as a Directed Acyclic Graph~(DAG).
Note that $\{(X_i, y_i)\}_i$ is given as the training data in the Few-shot setting, where $y_i \in Y$.

\section{Prompting Strategies}
\begin{figure}[t]
    \centering
    \small
    \begin{tcolorbox}[fontupper=\small]
    \begin{verbatim}
### Instructions
What area is this passage related to? You must select only one label
from ### Candidates and output the label following ### Answer.

### Candidates
Addiction
Algorithm design
...
network security

### Passage
{input data}

### Answer\end{verbatim}
    \end{tcolorbox}
    \caption{The prompt template for the DL strategy on the Web of Science dataset. \{input data\} area is replaced with actual input text.}
    \label{fig:DL_prompt}
\end{figure}

\begin{figure}[t]
    \centering
    \small
    \begin{tcolorbox}[fontupper=\small]
    \begin{verbatim}
### Instructions
What area is this passage related to? You must select only one
label from ### Candidates and output the label following ### Answer.
Candidate labels are given in a hierarchical structure in the
following form:

[1st depth label] > [2nd depth label]

### Candidates
Medical Sciences > Atopic Dermatitis
Medical Sciences > Alzheimer's Disease
...
Mechanical Engineering > computer-aided design

### Passage
{input data}

### Answer\end{verbatim}
    \end{tcolorbox}
    \caption{The prompt template for the DH strategy on the Web of Science dataset. \{input data\} area is replaced with actual input text.}
    \label{fig:DH_prompt}
\end{figure}
We conducted a comparative experiment on accuracy and cost using the following three prompting strategies: Direct Leaf Label Prediction Strategy~(DL), Direct Hierarchical Label Prediction Strategy~(DH), and Top-down Multi-step Hierarchical Label Prediction Strategy~(TMH). We explain each strategy in detail below.

\textit{Direct Leaf Label Prediction Strategy (DL)} select the corresponding label from the leaf nodes of the candidate label set $Y$ for each input text. The actual prompt template used is shown in Figure~\ref{fig:DL_prompt}. All leaf nodes are presented to the LLMs as candidate labels, and they are instructed to select one of them that matches the input text for output, as shown in Figure~\ref{fig:teaser_dl}.

\textit{Direct Hierarchical Label Prediction Strategy (DH)} causes the output to be a path on a set of candidate labels consisting of corresponding labels for each input text. The actual prompt template used is shown in Figure~\ref{fig:DH_prompt}. We input the path on the set of candidate labels in the form ``$\text{(1st depth)}>\text{(2nd depth)}>\cdots >\text{(leaf node)}$'' as candidate labels to the LLMs, as shown in Figure~\ref{fig:teaser_dh}.

In \textit{Top-down Multi-step Hierarchical Label Prediction Strategy (TMH)} estimate successively the labels for each depth by repeatedly selecting the most appropriate label for each level of the hierarchy, presenting the set of child labels of that label as candidate labels for the next step, and making label predictions, as shown in Figure~\ref{fig:teaser_tmh}. LLMs don't always return outputs that exactly match the candidate labels. Therefore, in this strategy, if the candidate label set includes the predicted label, we regarded the label with the smallest Levenshtein distance from the LLM’s output as the predicted label and presented its corresponding child labels as the candidate label set for the next depth level.
The prompts for each depth in this strategy are nearly identical to those shown in Figure~\ref{fig:DL_prompt}, with the only difference being that the Candidates section contains only the child labels of the predicted labels from the previous depth.
A key challenge in this strategy is that LLMs do not always return labels strictly from the provided candidate label set when predicting lower-level labels after predicting higher-level ones. To address this issue, if the predicted labels are not included in the candidate label set, the child labels of the candidate labels are identified as the closest matches using Levenshtein distance and are subsequently presented in the candidate label set at later stages.

\section{Experiments}
To investigate how the performance and cost of hierarchical text classification varies when various prompting strategies are used with black box large language models. The black box LLMs used in the experiments were gpt-4o-mini~(gpt-4o-mini-2024-07-18) provided by OpenAI. The codes are available at [Anonymous Link]\footnote{The code used in this study will be made publicly available after the paper is accepted.}.

\subsection{Setup}
In this experiment, we perform label prediction for each predefined strategy and dataset using zero-shot and few-shot prompting. For few-shot prompting, we randomly sample examples from the training data to construct the prompt. Specifically, we randomly select a specified number of examples from the training data of each dataset. The number of examples is determined based on predefined criteria for each dataset.

We use the gpt-4o-mini-2024-07-18 model, setting the temperature and top\_p parameters to 1.0. These values are chosen to maintain diversity in generated responses while ensuring a balanced level of randomness.

\subsubsection{Datasets}
\begin{figure*}[t]
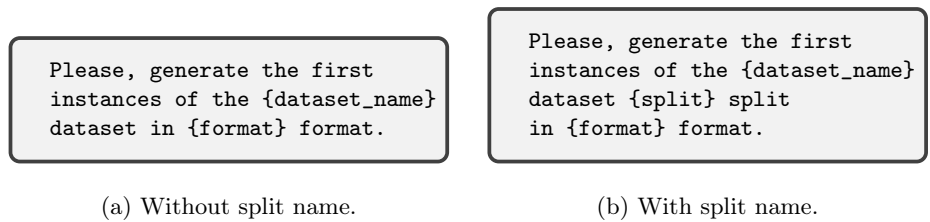

\begin{minipage}[b]{0.48\textwidth}
    \centering
    \begin{tcolorbox}
    \begin{verbatim}
Please, generate the first
instances of the {dataset_name}
dataset in {format} format.\end{verbatim}
    \end{tcolorbox}
    \subcaption{Without split name.}
    \label{fig:chatgpt_cheat_prompt_without_split_name}
    \end{minipage}
    \hspace{0.03\textwidth}
    \begin{minipage}[b]{0.48\textwidth}
    \centering
    \begin{tcolorbox}
    \begin{verbatim}
Please, generate the first
instances of the {dataset_name}
dataset {split} split
in {format} format.\end{verbatim}
    \end{tcolorbox}
    \subcaption{With split name.}
    \label{fig:chatgpt_cheat_prompt_with_split_name}    
    \end{minipage}
    \caption{The prompts of \textit{ChatGPT-Cheat?} to validate data contamination. \{dataset\_name\} is replaced with a target dataset name, \{split\} is replaced with a target split name, and \{format\} is replaced with a target data format type.}
\end{figure*}

\begin{table}[t]
\caption{The detail of Web of Science and Amazon Product Reviews dataset. The \textbf{\#(candidate labels)} part in the table represents the number of labels at each depth of the hierarchical classification.}
 \label{tb:dataset_details}
 \centering
 \renewcommand{\arraystretch}{1.2}
\begin{tabular}{l R R R R R}
\toprule
& \multicolumn{2}{c}{\textbf{\#(data)}} & \multicolumn{3}{c}{\textbf{\#(candidate labels)}} \\ \cmidrule(lr){2-3} \cmidrule(lr){4-6}
\textbf{dataset name} & \textbf{train} & \textbf{test} & \textbf{1st} &  \textbf{2nd} & \textbf{3rd} \\ \midrule
Web of Science & 1,250 & 1,800 & 7 & 136 & - \\
Amazon Product Reviews & 1,250 & 1,800 & 6 & 62 & 309 \\  \bottomrule
\end{tabular}
\end{table}

We conducted experiments using two datasets: Web of Science (WOS)\cite{hdltex} and Amazon Product Reviews (APR)\cite{aprdata}. Details of the dataset are provided in Table~\ref{tb:dataset_details}. To ensure the integrity of our evaluation, we performed data contamination checks using ChatGPT-Cheat?~\cite{chatgptcheat} for both datasets. We classified responses into four categories: contaminated (direct dataset reproduction), suspicious (output of characteristic attributes), safety-filtered (blocked output), and clean (no contamination). The parameters were set to temperature = 0 and max\_completion\_tokens = 500 for all models. Since WOS is originally in .xlsx format, we also tested contamination in .csv format, considering potential LLM training sources. As WOS lacks predefined train/valid/test splits, we used prompts~(in Figure~\ref{fig:chatgpt_cheat_prompt_without_split_name}) that do not reference specific dataset partitions. We tested both names for APR, called "Hierarchical Text Classification," another name to ensure thorough verification. Additionally, since APR is split into train and validation sets, we used prompts~(in Figure~\ref{fig:chatgpt_cheat_prompt_with_split_name}) explicitly mentioning these splits. The results of this data contamination check are summarized, where no contamination or suspicious cases were detected, confirming the validity of these datasets for evaluating LLM performance in hierarchical text classification.

For a more rigorous contamination assessment, we further employed Time-Travel-in-LLMs~\cite{golchin2024time} at the instance level. Based on this analysis, we selected 1,800 uncontaminated instances as the test set. The remaining data, after excluding these test instances, were used to construct the training set, from which we randomly sampled 1,250 instances as training data.

The WOS dataset~\cite{hdltex} is a collection of 46,985 published papers collected from the Web of Science. Abstracts, domains, and keywords are extracted from each article, and a hierarchical text classification dataset is constructed with abstracts as the input text, domains as first-depth labels, and keywords as second-depth labels.

The APR dataset~\cite{aprdata} is for the Review and Product categories collected by scraping from amazon.com and published on kaggle.com. The 40,000 records published as training data are labeled across three tiers, one for each tier. We use a subset of these 40,000 records as both training and test data in our experiments, following the method described earlier.


\subsubsection{Evaluation Metric}
We evaluate performance using accuracy. As LLMs do not necessarily output the labels in the set of labels shown as candidates as answers, the performance is underestimated if the accuracy of normal multi-class classification is applied based on whether or not there is perfect agreement. Therefore, text normalization processing is applied to both the output of LLMs and Ground Truth before evaluation. 
We remove some symbols and decapitalize text as part of the text normalization process.
We denote the accuracy value by $ACC_d$ in depth $d$. In addition, we calculated the accuracy of the child labels when the parent label matched the ground truth and denoted the accuracy value by $P(p_{d+1}^{True}|p_{d}^{True})$ when the parent label depth is $d$, and the child label depth is $d+1$.

For DL, no other than leaf labels are estimated, so the hierarchy above the leaf is estimated by tracing the parent labels of the leaf labels.

\subsubsection{Baseline Methods}
 To evaluate the performance of black box LLMs for HTC, we compare them with Hierarchy-aware Prompt Tuning for Hierarchical Text Classification (HPT)~\cite{hpt}, a non-LLM machine learning-based approach from conventional research. HPT is a hierarchical text classification method that leverages a transformer-based architecture while incorporating hierarchical label dependencies to improve classification accuracy.
We set batch\_size = 16 for the parameter settings while keeping all other parameters at their default values. We conducted the experiments using the official implementation available at \url{https://github.com/wzh9969/HPT}.

\subsection{Results}
\begin{table}[t]
 \caption{
 Results of the Web of Science dataset. Performance results for zero-shot and few-shot prompting across the three prompt strategies, along with comparisons to a machine learning model. The best-performing prompt strategy in each setting is highlighted in bold.
 }
 \label{tb:wos}
 \centering
  \renewcommand{\arraystretch}{1.2}
 \setlength{\tabcolsep}{6pt}
\begin{tabular}{lcccc}
\toprule
\textbf{Method} & \textbf{\#(Few Shot)} & $\mathbf{ACC_1}$ & $\mathbf{P(p_2^{True}|p_1^{True})}$ & $\mathbf{ACC_2}$ \\ \midrule
\multicolumn{5}{c}{\textbf{Machine Learning Model}} \\ \midrule
HPT &  & 0.826 & 0.655 & 0.571 \\ \midrule
\multicolumn{5}{c}{\textbf{Prompt Strategies}} \\ \midrule
DL  & 0                     & 0.677 & 0.581 & 0.393 \\
DL  & 1                     & 0.707 & 0.604 & 0.427 \\
DL  & 3                     & 0.708 & 0.620 & 0.439 \\
DL  & 5                     & \textbf{0.713} & 0.617 & \textbf{0.440} \\
DL  & 10                    & 0.712 & 0.605 & 0.431 \\
DL  & 20                    & 0.710 & 0.611 & 0.434 \\
DH  & 0                     & 0.627 & 0.601 & 0.401 \\
DH  & 1                     & 0.693 & 0.598 & 0.434 \\
DH  & 3                     & 0.688 & 0.579 & 0.417 \\
DH  & 5                     & 0.691 & 0.572 & 0.413 \\
DH  & 10                    & 0.688 & 0.567 & 0.407 \\
DH  & 20                    & 0.684 & 0.575 & 0.416 \\
TMH & 0                     & 0.616 & 0.652 & 0.405 \\
TMH & 1                     & 0.654 & 0.664 & 0.436 \\
TMH & 3                     & 0.652 & \textbf{0.665} & 0.434 \\
TMH & 5                     & 0.651 & 0.653 & 0.427 \\
TMH & 10                    & 0.656 & 0.657 & 0.433 \\
TMH & 20                    & 0.654 & 0.663 & 0.437 \\ \bottomrule
\end{tabular}
\end{table}

\begin{table*}[t]
 \caption{Results of Amazon Product Reviews dataset.
 Performance results for zero-shot and few-shot prompting across the three prompt strategies, along with comparisons to a machine learning model. The best-performing prompt strategy in each setting is highlighted in bold.}
 \label{tb:apr}
 \centering
   \renewcommand{\arraystretch}{1.2}
 \setlength{\tabcolsep}{2pt}
\begin{tabular}{lcccccc}
\toprule
        \textbf{Method} & \textbf{\#(Few Shot)} & $\mathbf{ACC_1}$ & $\mathbf{P(p_2^{True}|p_1^{True})}$ & $\mathbf{ACC_2}$ & $\mathbf{P(p_3^{True}|p_2^{True})}$ & $\mathbf{ACC_3}$ \\ \midrule
\multicolumn{7}{c}{\textbf{Machine Learning Model}} \\ \midrule
HPT &  & 0.823 & 0.657 & 0.556 & 0.641 & 0.377 \\ \midrule
\multicolumn{7}{c}{\textbf{Prompt Strategies}} \\ \midrule
DL  & 0                     & 0.637 & 0.561 & 0.357 & 0.720  & 0.257 \\
DL  & 1                     & 0.667 & 0.629 & 0.419 & 0.768 & 0.322 \\
DL  & 3                     & 0.693 & 0.675 & 0.468 & 0.785 & 0.367 \\
DL  & 5                     & 0.690  & 0.688 & 0.474 & 0.783 & 0.372 \\
DL  & 10                    & 0.701 & 0.679 & 0.476 & 0.788 & 0.375 \\
DL  & 20                    & 0.709 & 0.707 & 0.502 & 0.781 & 0.392 \\
DH  & 0                     & 0.817 & 0.718 & 0.591 & 0.782 & 0.491 \\
DH  & 1                     & 0.854 & 0.718 & 0.616 & 0.784 & 0.510  \\
DH  & 3                     & 0.862 & 0.732 & 0.633 & 0.770  & 0.507 \\
DH  & 5                     & \textbf{0.868} & 0.733 & 0.640  & 0.769 & 0.517 \\
DH  & 10                    & 0.867 & \textbf{0.744} & \textbf{0.649} & 0.769 & 0.521 \\
DH  & 20                    & 0.854 & \textbf{0.744} & 0.646 & 0.796 & \textbf{0.532} \\
TMH & 0                     & 0.847 & 0.68  & 0.576 & 0.754 & 0.436 \\
TMH & 1                     & 0.824 & 0.679 & 0.560  & 0.783 & 0.440  \\
TMH & 3                     & 0.828 & 0.673 & 0.558 & 0.793 & 0.442 \\
TMH & 5                     & 0.825 & 0.678 & 0.560  & 0.811 & 0.455 \\
TMH & 10                    & 0.836 & 0.681 & 0.570  & 0.842 & 0.481 \\
TMH & 20                    & 0.828 & 0.691 & 0.573 & \textbf{0.853} & 0.490 \\ \bottomrule
  \end{tabular}
\end{table*}
We present the experimental results of hierarchical classification using three different prompt strategies with a black box LLM. The evaluation is performed on two datasets: the Web of Science dataset and the Amazon Product Reviews dataset. The results are compared against a machine learning model (HPT) to assess the effectiveness of LLM-based prompting strategies in few-shot and zero-shot settings.
Table~\ref{tb:wos} shows the performance of different prompt strategies on the Web of Science dataset. The performance is measured using three metrics: $ACC_1$, $P(p_2^{True} | p_1^{True})$, and $ACC_2$. Among the three prompt strategies, DL with 5-shot prompting achieves the highest $ACC_1$ (0.713) and $ACC_2$ (0.440), demonstrating the strongest classification performance at both levels of the hierarchy. In terms of $P(p_2^{True} | p_1^{True})$, which measures the conditional probability of correctly predicting the second-level class given a correct first-level prediction, the TMH strategy with 3-shot prompting achieves the highest value (0.665), outperforming other settings. The machine learning model (HPT) outperforms all LLM-based approaches, with $ACC_1$ = 0.826, $P(p_2^{True} | p_1^{True})$ = 0.655, and $ACC_2$ = 0.571. Zero-shot prompting generally yields lower performance compared to few-shot settings, emphasizing the necessity of in-context learning to improve classification results.
Table~\ref{tb:apr} presents the results for the Amazon Product Reviews dataset. This dataset involves a three-level hierarchical classification task, and we evaluate performance using five metrics: $ACC_1$, $P(p_2^{True} | p_1^{True})$, $ACC_2$, $P(p_3^{True} | p_2^{True})$, and $ACC_3$. The DH prompt strategy consistently outperforms DL and TMH, particularly in few-shot settings. Specifically, DH with 5-shot prompting achieves the highest $ACC_1$ (0.868) and $ACC_2$ (0.640). The highest $P(p_2^{True} | p_1^{True})$ (0.744) is observed in DH with 10-shot and 20-shot prompting, showing the effectiveness of deeper hierarchical prompting. For the final level classification ($ACC_3$), the best performance (0.532) is achieved by DH with 20-shot prompting. The TMH strategy achieves the highest $P(p_3^{True} | p_2^{True})$ (0.853) in the 20-shot setting, suggesting its advantage in preserving classification consistency at deeper hierarchical levels. As in the Web of Science dataset, the machine learning model (HPT) generally outperforms LLM-based prompting strategies, achieving $ACC_1$ = 0.823, $ACC_2$ = 0.556, and $ACC_3$ = 0.377.

Overall, the results demonstrate that few-shot prompting significantly improves performance over zero-shot prompting across all strategies. The effectiveness of few-shot prompting strategies varies depending on the dataset. In the Web of Science dataset, the machine learning model maintains a clear advantage over all prompting strategies. However, in the Amazon Product Reviews dataset, where the label structure is more complex and the amount of training data is relatively limited, certain few-shot prompting strategies, particularly DH and TMH, achieve performance comparable to the machine learning model. These findings suggest that selecting an appropriate prompt strategy and increasing the number of examples in the prompt can significantly enhance the performance of LLMs for hierarchical classification tasks, particularly in scenarios with constrained training data.

\subsection{Cost Analysis}
\begin{table}[t]
 \caption{Average number of prompt tokens (input tokens) on the upper part and completion tokens (output tokens) on the lower part of the table.}
 \label{tb:cost}
    \centering
       \renewcommand{\arraystretch}{1}
 \setlength{\tabcolsep}{6pt}
    \begin{tabular}{llrrrrrr}
\toprule
 & & \multicolumn{6}{c}{\textbf{\#(few shot examples)}}  \\ \cmidrule{3-8}
\textbf{dataset} &
  \textbf{prompt} &
  \textbf{0} & \textbf{1} & \textbf{3} & \textbf{5} & \textbf{10} & \textbf{20} \\ \midrule
\multicolumn{8}{c}{\textbf{prompt tokens}} \\ \midrule
WOS &
  DL & 833.33 & 1105.00 & 1662.39 & 2210.69 & 3594.35 & 6326.98 \\
                       & DH  & 1249.33 & 1523.39 & 2080.72 & 2642.91 & 4034.88 & 6822.23  \\ 
                       & TMH & 783.70   & 1305.11 & 2389.67 & 3491.28 & 6250.63 & 11755.44 \\ 
APR &
  DL &
  1337.16 &
  1440.54 &
  1653.19 &
  1866.96 &
  2377.60 &
  3424.61 \\
                       & DH  & 3354.16 & 3465.7                       & 3689.27                      & 3912.13 & 4460.17 & 5574.73  \\
                       & TMH & 511.23  & 828.81                       & 1444.18                      & 2057.71 & 3559.82 & 6472.83  \\ \toprule
\multicolumn{8}{c}{\textbf{completion tokens}} \\ \midrule
WOS         & DL  & 4.47    & 3.90                          & 3.64                         & 3.67    & 3.47    & 3.83     \\
                       & DH  & 6.30     & 6.23                         & 6.21                         & 6.22    & 6.23    & 6.33     \\
                       & TMH & 7.51    & 6.81                         & 7.07                         & 6.78    & 6.95    & 7.03     \\
APR & DL  & 4.49    & 3.83                         & 3.92                         & 4.03    & 4.08    & 4.11     \\
                       & DH  & 9.99    & 10.06                        & 10.10                         & 10.08   & 10.10   & 10.07    \\
                       & TMH & 12.58   & 11.32                        & 11.45                        & 11.25   & 11.51   & 11.33  \\ \bottomrule
\end{tabular}
\end{table}
Here, we analyze the computational cost in terms of the number of input tokens (prompt tokens) and output tokens (completion tokens) required for our approach under different few-shot settings. Table~\ref{tb:cost} presents the average number of tokens used across different datasets and prompt configurations.

The number of prompt tokens increases with more few-shot examples, significantly impacting computational cost. In the WOS dataset, the DL prompt grows from 833.33 tokens (zero-shot) to 6326.98 tokens (20-shot), while the TMH prompt reaches 11755.44 tokens. Similarly, in the APR dataset, the DH prompt expands from 3354.16 to 5574.73 tokens. In contrast, completion tokens remain stable across settings, fluctuating only slightly. This suggests that prompt tokens are the primary cost factor rather than output tokens.

Each prompt type (DL, DH, and TMH) exhibits distinct cost characteristics. The DH prompt consistently requires the highest number of prompt tokens, indicating that it demands more extensive context or detailed information, leading to higher computational costs. In contrast, the DL prompt shows a more moderate increase in token usage, suggesting it balances brevity and informativeness effectively. TMH prompts, while starting with fewer tokens, scale up dramatically with increasing few-shot examples, making them highly sensitive to the number of examples used.

Given these differences, a cost-effective prompt selection strategy should account for the characteristics of each prompt type. DH achieves high accuracy when the label hierarchy is deep and the candidate set is large, as it leverages hierarchical structure effectively, but it generally incurs higher costs. For TMH prompts, limiting the number of few-shot examples is essential to avoid excessive token consumption. DL prompts offer a more predictable cost-performance tradeoff compared to DH and TMH prompts but still require careful token management. A well-optimized prompt selection strategy, informed by these insights, can balance model effectiveness and computational cost, ensuring efficient deployment of large language models.

\section{Conclusion}
This study explored the use of black box Large Language Models (LLMs) for Hierarchical Text Classification (HTC), aiming to address the challenges of data scarcity and model complexity. By employing prompting strategies instead of model training, we sought to achieve high accuracy with minimal labeled data and computational overhead. Three different prompting strategies were evaluated: Direct Leaf Label Prediction (DL), Direct Hierarchical Label Prediction (DH), and Top-down Multi-step Hierarchical Label Prediction (TMH), using both zero-shot and few-shot settings.

The experimental results demonstrate that LLM-based prompting strategies can achieve performance comparable to traditional machine learning models, depending on the dataset and hierarchy depth. In the Web of Science dataset, the machine learning model exhibited the best overall performance. However, in the Amazon Product Reviews dataset, where the label structure is more complex and the number of training samples is relatively limited, certain few-shot prompting strategies, particularly DH and TMH, achieved accuracy close to that of the machine learning model. This suggests that LLMs, when appropriately prompted, can serve as an effective alternative to traditional machine learning methods for HTC, particularly in low-resource scenarios.

Furthermore, the results highlight a trade-off between accuracy and computational cost. Few-shot prompting significantly improved classification performance across both datasets, often narrowing the gap between LLM-based and machine learning-based approaches. However, strategies such as DH, while achieving the highest classification accuracy, also incurred higher API costs as hierarchy depth increased. These findings indicate that prompting-based HTC with LLMs is a viable alternative to machine learning models, provided that computational cost is carefully managed.

This study has several limitations. Our analysis was limited to OpenAI’s GPT-4o mini and datasets with only two- to three-depth hierarchies. While deeper hierarchies may benefit from DH, further experiments on more complex datasets are needed. Additionally, restricting the study to black box LLMs limits our findings; future work should include other black box LLMs and fine-tuned white box LLMs to better understand cost-effectiveness and performance trade-offs.

\bibliographystyle{IEEEtran}
\bibliography{references}

\begin{thebibliography}{10}
\providecommand{\url}[1]{#1}
\csname url@samestyle\endcsname
\providecommand{\newblock}{\relax}
\providecommand{\bibinfo}[2]{#2}
\providecommand{\BIBentrySTDinterwordspacing}{\spaceskip=0pt\relax}
\providecommand{\BIBentryALTinterwordstretchfactor}{4}
\providecommand{\BIBentryALTinterwordspacing}{\spaceskip=\fontdimen2\font plus
\BIBentryALTinterwordstretchfactor\fontdimen3\font minus \fontdimen4\font\relax}
\providecommand{\BIBforeignlanguage}[2]{{%
\expandafter\ifx\csname l@#1\endcsname\relax
\typeout{** WARNING: IEEEtran.bst: No hyphenation pattern has been}%
\typeout{** loaded for the language `#1'. Using the pattern for}%
\typeout{** the default language instead.}%
\else
\language=\csname l@#1\endcsname
\fi
#2}}
\providecommand{\BIBdecl}{\relax}
\BIBdecl

\bibitem{hdltex}
K.~Kowsari, D.~E. Brown, M.~Heidarysafa, K.~Jafari~Meimandi, M.~S. Gerber, and L.~E. Barnes, ``Hdltex: Hierarchical deep learning for text classification,'' in \emph{2017 16th IEEE International Conference on Machine Learning and Applications}, 2017.

\bibitem{Gargiulo2019}
F.~Gargiulo, S.~Silvestri, M.~Ciampi, and G.~De~Pietro, ``Deep neural network for hierarchical extreme multi-label text classification,'' \emph{Applied Soft Computing}, 2019.

\bibitem{MOSKOVITCH2006177}
R.~Moskovitch, S.~Cohen-Kashi, U.~Dror, I.~Levy, A.~Maimon, and Y.~Shahar, ``Multiple hierarchical classification of free-text clinical guidelines,'' \emph{Artificial Intelligence in Medicine}, 2006.

\bibitem{Bhambhoria2023}
R.~Bhambhoria, L.~Chen, and X.~Zhu, ``A simple and effective framework for strict {Zero-Shot} hierarchical classification,'' in \emph{Proceedings of the 61st Annual Meeting of the Association for Computational Linguistics}.\hskip 1em plus 0.5em minus 0.4em\relax Association for Computational Linguistics, 2023.

\bibitem{gpt3}
T.~Brown, B.~Mann, N.~Ryder, M.~Subbiah, J.~D. Kaplan, P.~Dhariwal, A.~Neelakantan, P.~Shyam, G.~Sastry, A.~Askell, S.~Agarwal, A.~Herbert-Voss, G.~Krueger, T.~Henighan, R.~Child, A.~Ramesh, D.~Ziegler, J.~Wu, C.~Winter, C.~Hesse, M.~Chen, E.~Sigler, M.~Litwin, S.~Gray, B.~Chess, J.~Clark, C.~Berner, S.~McCandlish, A.~Radford, I.~Sutskever, and D.~Amodei, ``Language models are few-shot learners,'' in \emph{Advances in Neural Information Processing Systems}, 2020.

\bibitem{llama2}
H.~Touvron, L.~Martin, K.~Stone, P.~Albert, A.~Almahairi, Y.~Babaei, N.~Bashlykov, S.~Batra, P.~Bhargava, S.~Bhosale, D.~Bikel, L.~Blecher, C.~C. Ferrer, M.~Chen, G.~Cucurull, D.~Esiobu, J.~Fernandes, J.~Fu, W.~Fu, B.~Fuller, C.~Gao, V.~Goswami, N.~Goyal, A.~Hartshorn, S.~Hosseini, R.~Hou, H.~Inan, M.~Kardas, V.~Kerkez, M.~Khabsa, I.~Kloumann, A.~Korenev, P.~S. Koura, M.-A. Lachaux, T.~Lavril, J.~Lee, D.~Liskovich, Y.~Lu, Y.~Mao, X.~Martinet, T.~Mihaylov, P.~Mishra, I.~Molybog, Y.~Nie, A.~Poulton, J.~Reizenstein, R.~Rungta, K.~Saladi, A.~Schelten, R.~Silva, E.~M. Smith, R.~Subramanian, X.~E. Tan, B.~Tang, R.~Taylor, A.~Williams, J.~X. Kuan, P.~Xu, Z.~Yan, I.~Zarov, Y.~Zhang, A.~Fan, M.~Kambadur, S.~Narang, A.~Rodriguez, R.~Stojnic, S.~Edunov, and T.~Scialom, ``Llama 2: Open foundation and fine-tuned chat models,'' 2023.

\bibitem{zeroshottextclassifiers}
Z.~Wang, Y.~Pang, and Y.~Lin, ``Large language models are {Zero-Shot} text classifiers,'' 2023.

\bibitem{tabllm}
S.~Hegselmann, A.~Buendia, H.~Lang, M.~Agrawal, X.~Jiang, and D.~Sontag, ``Tabllm: Few-shot classification of tabular data with large language models,'' in \emph{Proceedings of The 26th International Conference on Artificial Intelligence and Statistics}, 2023.

\bibitem{hsvm}
N.~Cesa-bianchi, C.~Gentile, A.~Tironi, and L.~Zaniboni, ``Incremental algorithms for hierarchical classification,'' in \emph{Advances in Neural Information Processing Systems}, 2004.

\bibitem{bsvm}
N.~Cesa-Bianchi, C.~Gentile, and L.~Zaniboni, ``Hierarchical classification: combining bayes with svm,'' in \emph{Proceedings of the 23rd International Conference on Machine Learning}, 2006.

\bibitem{hpt}
Z.~Wang, P.~Wang, T.~Liu, B.~Lin, Y.~Cao, Z.~Sui, and H.~Wang, ``{HPT}: Hierarchy-aware prompt tuning for hierarchical text classification,'' in \emph{Proceedings of the 2022 Conference on Empirical Methods in Natural Language Processing}.\hskip 1em plus 0.5em minus 0.4em\relax Association for Computational Linguistics, 2022.

\bibitem{hierverb}
K.~Ji, Y.~Lian, J.~Gao, and B.~Wang, ``Hierarchical verbalizer for few-shot hierarchical text classification,'' in \emph{Proceedings of the 61st Annual Meeting of the Association for Computational Linguistics (Volume 1: Long Papers)}, A.~Rogers, J.~Boyd-Graber, and N.~Okazaki, Eds.\hskip 1em plus 0.5em minus 0.4em\relax Association for Computational Linguistics, 2023.

\bibitem{llmrec}
Y.~Chang, X.~Wang, J.~Wang, Y.~Wu, L.~Yang, K.~Zhu, H.~Chen, X.~Yi, C.~Wang, Y.~Wang, W.~Ye, Y.~Zhang, Y.~Chang, P.~S. Yu, Q.~Yang, and X.~Xie, ``A survey on evaluation of large language models,'' \emph{ACM Transactions on Intelligent Systems and Technology}, 2024.

\bibitem{teleclass}
Y.~Zhang, R.~Yang, X.~Xu, J.~Xiao, J.~Shen, and J.~Han, ``Teleclass: Taxonomy enrichment and llm-enhanced hierarchical text classification with minimal supervision,'' 2024.

\bibitem{chen-etal-2024-retrieval-style}
H.~Chen, Y.~Zhao, Z.~Chen, M.~Wang, L.~Li, M.~Zhang, and M.~Zhang, ``Retrieval-style in-context learning for few-shot hierarchical text classification,'' \emph{Transactions of the Association for Computational Linguistics}, vol.~12, 2024.

\bibitem{10825412}
F.~Schmidt, K.~Hammerfald, H.~H. Jahren, A.~H. Payberah, and V.~Vlassov, ``{ Single-pass Hierarchical Text Classification with Large Language Models },'' in \emph{2024 IEEE International Conference on Big Data (BigData)}.\hskip 1em plus 0.5em minus 0.4em\relax Los Alamitos, CA, USA: IEEE Computer Society, Dec. 2024.

\bibitem{aprdata}
\BIBentryALTinterwordspacing
Y.~Kashnitsky, ``Hierarchical text classification,'' 2020. [Online]. Available: \url{https://www.kaggle.com/dsv/1054619}
\BIBentrySTDinterwordspacing

\bibitem{chatgptcheat}
\BIBentryALTinterwordspacing
O.~Sainz, J.~A. Campos, I.~García-Ferrero, J.~Etxaniz, and E.~Agirre, ``Did chatgpt cheat on your test?'' 2023. [Online]. Available: \url{https://hitz-zentroa.github.io/lm-contamination/blog/}
\BIBentrySTDinterwordspacing

\bibitem{golchin2024time}
S.~Golchin and M.~Surdeanu, ``Time travel in {LLM}s: Tracing data contamination in large language models,'' in \emph{The Twelfth International Conference on Learning Representations}, 2024.

\end{thebibliography}

\newpage
\appendix

\end{document}